\documentclass[10pt,twocolumn,letterpaper]{article}

\usepackage[pagenumbers]{cvpr} 

\usepackage[dvipsnames]{xcolor}

\definecolor{cvprblue}{rgb}{0.21,0.49,0.74}
\usepackage[pagebackref,breaklinks,colorlinks,citecolor=cvprblue]{hyperref}

\newcommand{\OURS}{NViST}

\def\paperID{6582} 
\def\confName{CVPR}
\def\confYear{2024}

\title{\OURS{}: In the Wild New View Synthesis from a Single Image with Transformers}

\author{
Wonbong Jang \qquad 
Lourdes Agapito \\
Department of Computer Science\\
University College London\\
{\tt\small
    \{ucabwja,l.agapito\}@ucl.ac.uk
}
}

\begin{document}

\twocolumn[{%
  \renewcommand\twocolumn[1][]{#1}%
\maketitle
\begin{center}
  \newcommand{\teaserwidth}{\textwidth}
  \centerline{
    \includegraphics[width=\teaserwidth]{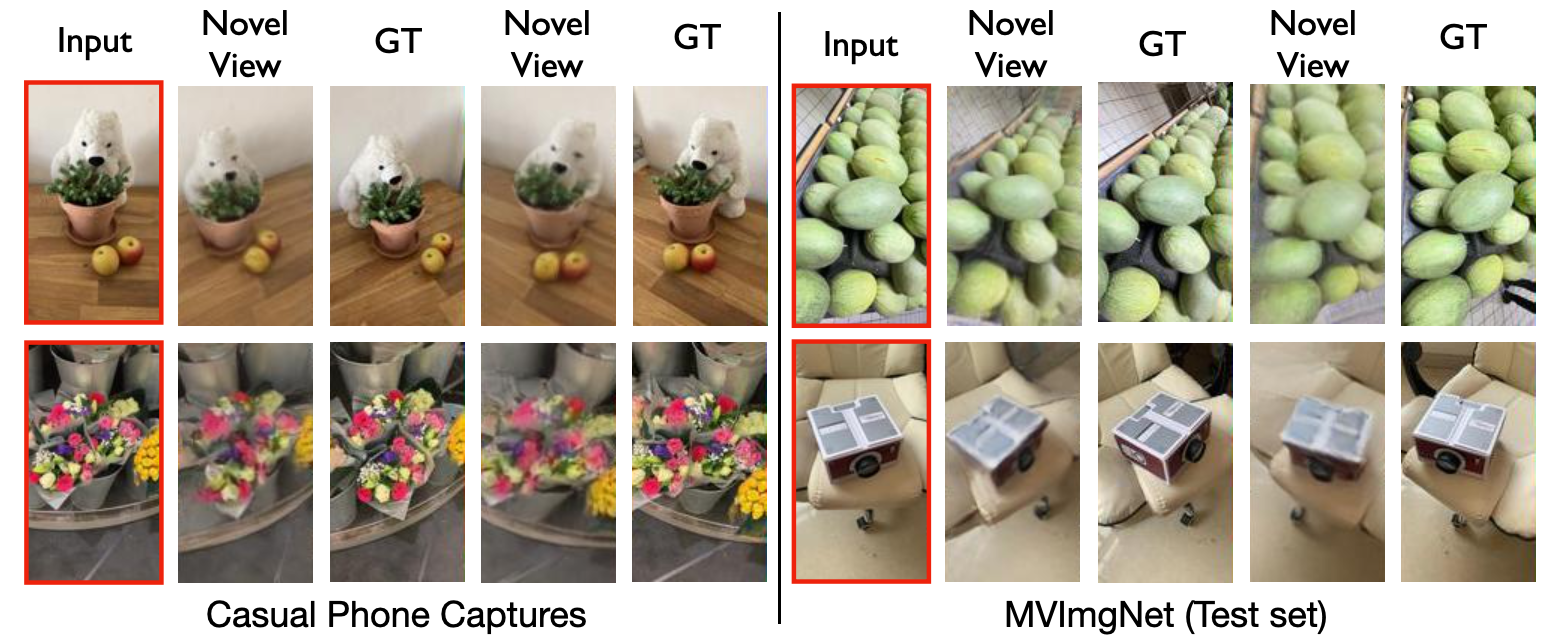}
    }
    \captionof{figure}{We introduce \OURS{}, a transformer-based architecture that enables synthesis from novel viewpoints given a single in the wild input image. We test our model not only on held-out scenes of MVImgNet, a large-scale dataset of casually captured videos of hundreds of object categories (Right) but also on out-of-distribution challenging phone-captured scenes (Left).}
  \label{fig:teaser}
 \end{center}
}]

\setcounter{page}{1}
\begin{abstract}

We propose \OURS{}, a transformer-based model for efficient and generalizable novel-view synthesis from a single image for real-world scenes. In contrast to many methods that are trained on synthetic data, object-centred scenarios, or in a category-specific manner, \OURS{} is trained on MVImgNet, a large-scale dataset of casually-captured real-world videos of hundreds of object categories with diverse backgrounds.
\OURS{} transforms image inputs directly into a radiance field, conditioned on camera parameters via adaptive layer normalisation. In practice, \OURS{} exploits fine-tuned masked autoencoder (MAE) features and translates them to 3D output tokens via cross-attention, while addressing occlusions with self-attention. 
To move away from object-centred datasets and enable full scene synthesis, \OURS{} adopts a 6-DOF camera pose model and only requires relative pose, dropping the need for canonicalization of the training data, which removes a substantial barrier to it being used on casually captured datasets. 
We show results on unseen objects and categories from MVImgNet and even generalization to casual phone captures. We conduct qualitative and quantitative evaluations on MVImgNet and ShapeNet to show that our model represents a step forward towards enabling true in-the-wild generalizable novel-view synthesis from a single image. Project webpage: \url{https://wbjang.github.io/nvist_webpage}.

\end{abstract}

\section{Introduction}

Learning 3D scene representations from RGB images for novel view synthesis or 3D modelling remains a pivotal challenge for the computer vision and graphics communities. 
Traditional approaches, such as structure from motion (SfM) and multiview stereo pipelines, endeavor to optimize the 3D scene from RGB images directly by leveraging geometric and photometric consistency. 
The advent of Neural Radiance Fields (NeRF)~\cite{mildenhall2020nerf} and its subsequent developments has marked a significant step forward by encoding the entire 3D scene within the weights of a neural network (or feature grid), from RGB images only.
Although NeRF requires more than dozens of images for training, efforts have been undertaken to generalize NeRF across multiple scenes by taking a single image as an input~\cite{yu2021pixelnerf, wang2021ibrnet, Henzler_2021_CVPR, jang2021codenerf, chen2021mvsnerf, lin2023visionnerf, muller2022autorf, rebain2022lolnerf}. 
Nonetheless, generalizing NeRF-based models to multiple real-world scenes remains a challenge due to scale ambiguities, scene misalignments and diverse backgrounds.
The huge success of 2D latent diffusion models~\cite{rombach2022high} has  sparked interest in 3D diffusion models.
One trend is to make diffusion models 3D-aware by fine-tuning a pre-trained 2D diffusion model. However, so far these approaches are trained on centered objects, masked inputs, do not deal with backgrounds, and assume a simplified camera model 
(3-DOF)~\cite{liu2023zero,liu2023one2345, qian2023magic123, shi2023zero123}. 
Other approaches~\cite{gu2023nerfdiff,chan2023generative,tewari2023forwarddiffusion,chen2023single} build the diffusion model on top of volume rendering in 3D space, but they are computationally expensive and slow to sample from.

Many recent breakthroughs in computer vision can be attributed to the emergence of very large datasets paired with the transformer architecture.
For instance, MiDAS~\cite{Ranftl2022} 
exploits  multiple diverse datasets, leading to robust performance in zero-shot depth estimation from a single RGB image 
while SAM~\cite{kirillov2023segment} demonstrates that an extensive dataset coupled with a pretrained MAE and prompt engineering can significantly improve performance in segmentation.
However, in 3D computer vision, scaling up real-world datasets has not been as straightforward. Synthetic datasets like ShapeNet~\cite{chang2015shapenet} or Objaverse~\cite{deitke2023objaverse} have helped to promote progress, but there is a large domain gap.

The recent release of real-world large-scale multiview datasets such as Co3D~\cite{reizenstein2021common} or MVImgNet~\cite{yu2023mvimgnet}, coupled with the availability of robust SfM tools such as COLMAP~\cite{schoenberger2016mvs, schoenberger2016sfm} or ORB-SLAM~\cite{mur2015orb} to enable camera pose estimation, has opened the door to large-scale training for new view synthesis. 
However, significant challenges remain to train a scene-level new view synthesis model on such real-world, large scale datasets due to the huge diversity of objects, categories, scene scales, backgrounds, and scene alignment issues.
Motivated by this we propose \OURS{}, a transformer-based architecture trained on a large-scale dataset to enable in-the-wild new view synthesis (NVS) from a single image. We exploit a subset of MVImgNet~\cite{yu2023mvimgnet} which has one order of magnitude more categories and $\times$2 more objects than Co3D~\cite{reizenstein2021common}. Our contributions can be summed up as follows:

\begin{itemize}
\item \OURS{} can model general real-world scenes, including backgrounds, only requiring relative pose during training.
\item Our novel decoder maps MAE features to 3D output tokens via cross-attention, conditions on camera parameters via adaptive layer normalisation and addresses occlusions with self-attention.
\item Our qualitative and quantitative evaluations on MVImgNet test sequences show good performance on challenging real-world scenes.  
\item We demonstrate good generalization results on a zero-shot new-view synthesis task, on phone-captured scenes.

\end{itemize}

\begin{figure*}
\vspace{-0.5cm}
  \centering
  \includegraphics[width=\textwidth]{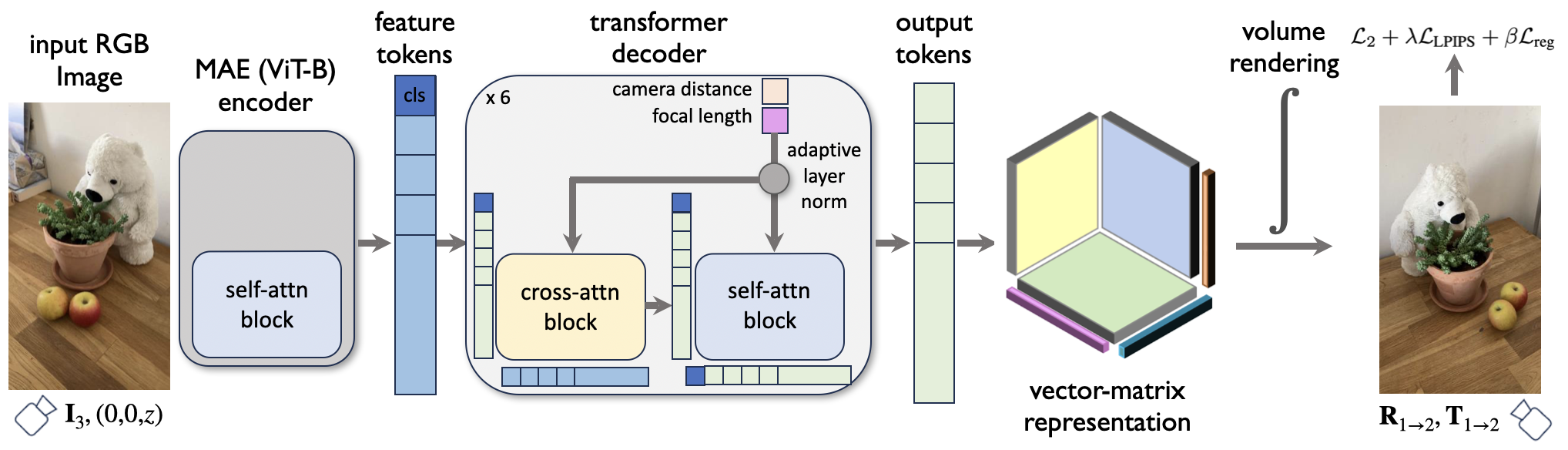}
  \caption{\textbf{Architecture.} \OURS{} is a feed-forward transformer-based model that takes a single in-the-wild image as input, and renders a novel view. The encoder, a finetuned Masked Autoencoder (MAE), generates feature tokens, which are translated to output tokens via cross-attention by 
  our novel decoder, conditioned on normalised focal length and camera distance via adaptive layer normalisation. 
  Self-attention blocks allow reasoning about occlusions. Output tokens are reshaped into a vector-matrix representation 
  that is used for volume rendering. 
  \OURS{} is trained end-to-end via a balance of losses: photometric $L_2$, perceptual $L_{\text{LPIPS}}$, and a distortion-based regulariser $L_{\text{reg}}$.
  } 
  \label{fig:overall}
\end{figure*}

\section{Related Work}

\textbf{Transformer, ViT and MAE:} The transformer~\cite{vaswani2017attention}, a feed-forward, scalable, attention-based architecture, has brought a revolution to the field of natural language understanding. 
Inspired by it, the vision transformer (ViT) ~\cite{dosovitskiy2020image} uses image patches as tokens 
to achieve performance levels comparable to those of CNNs~\cite{he2016deep, simonyan2014very} in many computer vision tasks. 
While the ViT is trained in a supervised way, masked autoencoders (MAE)~\cite{he2022masked} 
can be trained in a self-supervised way by randomly masking and in-painting patches, and be further fine-tuned on specific tasks. 

\noindent\textbf{Neural Implicit Representations: } Neural implicit representations aim to learn a 3D representation without direct 3D supervision using neural networks. 
They have been employed for various tasks, including depth prediction~\cite{niemeyer2020differentiable} and scene representation through ray marching and coordinate-based MLPs~\cite{sitzmann2019scene}.
\textit{Mildenhall et al.}~\cite{mildenhall2020nerf} proposed Neural Radiance Fields (NeRF), which integrates coordinate-based MLPs, positional encoding, and volume rendering to encode a scene in the weights of neural network. 
Upon optimisation, novel views can be rendered with impressively high quality. 
Beyond novel view synthesis, NeRF has found utility in a diverse range of computer vision tasks such as segmentation~\cite{zhi2021place, vora2021nesf, fu2022panoptic, cen2023segment}, surface reconstruction~\cite{oechsle2021unisurf, wang2021neus, wang2022go, mohamed2023dynamicsurf, Cai2022NDR}, and camera registration~\cite{lin2021barf,jeong2021self,chng2022gaussian, yen2021inerf, lin2022parallel, chen2022local, wang2021nerf,liu2023robust}.

\noindent\textbf{Grid-based representations:} A limitation of the original NeRF method is its lengthy training time. 
As an alternative to coordinate-based MLPs, grid-based approaches~\cite{yu2021plenoxels, sun2022direct, muller2022instant, chen2022tensorf, fridovich2023k, tilted2023} have been proposed to expedite training.  
TensoRF~\cite{chen2022tensorf} proposed the vector-matrix (VM) representation as an efficient and compact way to represent the 3D grid.
In the context of 3D-aware Generative Adversarial Networks (GANs), EG3D~\cite{chan2022efficient} introduced triplanes by projecting 3D features into three different planes. 
Several approaches use the triplane representation to learn to generate 3d implicit representations from ImageNet~\cite{3dgp,sargent2023vq3d,Schwarz2024ICLR,reddy2024g3dr}.

\noindent\textbf{Learning multiple scenes using NeRF: } Generalising NeRFs to multiple scenes remains a challenging problem. 
Several methods associate 2D features with the target views~\cite{yu2021pixelnerf, wang2021ibrnet, chen2021mvsnerf,reizenstein2021common, Henzler_2021_CVPR, trevithick2021grf,chen2023matchnerf, irshad2023neo360}, while others condition the network on latent vectors with a shared MLP across the dataset~\cite{Gafni_2021_CVPR, jang2021codenerf, muller2022autorf,rebain2022lolnerf,athar2022rignerf,hong2022headnerf}. 
Adding an adversarial loss to NeRF leads to 3D-aware GANs, which allow consistent rendering from different viewpoints~\cite{chanmonteiro2020pi-GAN, chan2022efficient, schwarz2020graf,le2023stylemorph,cai2022pix2nerf}.
Most approaches that use a single input image require aligned datasets like ShapeNet~\cite{chang2015shapenet, sitzmann2019scene} or FFHQ~\cite{karras2019ASG}, and fail to deal with scale ambiguities or diverse backgrounds.

\noindent\textbf{Diffusion for NVS: } Latent diffusion~\cite{rombach2022high} and its open-source release Stable Diffusion have transformed the field of image generation. 
However, applying diffusion models to learn 3D implicit representations is not straightforward as there is no access to 3D ground-truth.
3DiM~\cite{watson2022novel} proposes a pose-conditioned image-to-image approach.
Dreamfusion~\cite{poole2022dreamfusion} introduced score distillation sampling (SDS) to train NeRF with a 2D diffusion model. SDS has been used in many follow-up works~\cite{melas2023realfusion, shi2023mvdream, tang2023dreamgaussian, wang2023morpheus, wang2023prolificdreamer}. 
Fine-tuning Stable Diffusion on a large-scale synthetic dataset~\cite{deitke2023objaverse} allows diffusion models to be 3D-aware such as Zero-1-to-3 and its follow-ups~\cite{liu2023zero,qian2023magic123,liu2023one2345,shi2023zero123}.
Other approaches denoise directly in 3D and supervise the model in 2D space after rendering, ~\cite{karnewar2023holodiffusion, szymanowicz23viewset_diffusion,anciukevicius24iclr,gu2023nerfdiff, chan2023generative, tewari2023forwarddiffusion, anciukevivcius2023renderdiffusion}, use the 2D diffusion model as a prior~\cite{deng2023nerdi}, or optimise NeRF jointly and regard it as the ground truth~\cite{chen2023single}.
Similar to latent diffusion~\cite{kim2023nfldm, Schwarz2024ICLR, bautista2022gaudi} adopt 2-stage training.

\noindent\textbf{3D Representation with Transformers: } Geometry-free methods employing transformer architectures have been explored as seen in  ~\cite{kulhanek2022viewformer, sajjadi2022scene, sajjadi2022osrt, sajjadi2022rust}.
Others build NeRF representations using transformers~\cite{lin2023visionnerf, sargent2023vq3d, shen2023gina, wu2023multiview, jun2023shap}.
The concurrent work LRM~\cite{hong2023lrm} extracts image features through DINO~\cite{caron2021emerging}, and refines learnable positional embeddings via attention mechanisms. 
Unlike ~\OURS{}, LRM focuses on  object-centric scenes without background and employs triplanes instead of vector-matrix representation. 
\section{Methodology}

\OURS{} is a feed-forward conditional encoder-decoder model built upon transformers to predict a radiance field from a single image to enable synthesis from novel viewpoints. As Figure \ref{fig:overall} shows, 
our architecture is structured into three key components: two transformer-based modules (encoder and decoder) and a NeRF renderer, which is composed of a shallow multi-layer perceptron (MLP) and a differentiable volume rendering module. 

\subsection{Encoder}

\begin{figure}[tb]
  \centering
  \includegraphics[width=0.48\textwidth]{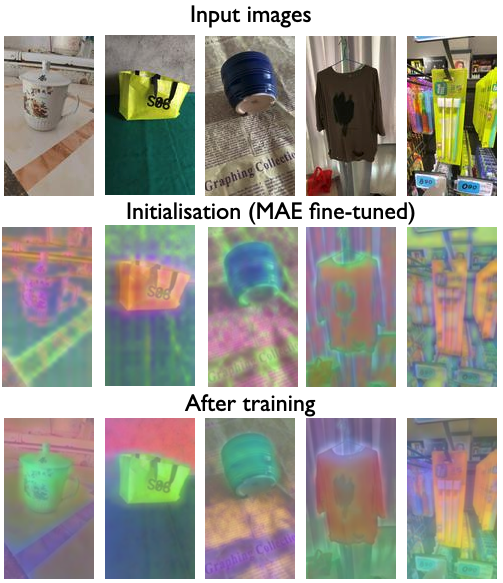}
    \caption{\textbf{Encoder Output Visualisation:}  (Top) Input images. (Middle) Features from a fine-tuned MAE, which serve as initialisation to our encoder. (Bottom) Features after end-to-end training. Features shown after reducing to $3$ dimensions with PCA. Optimised features appear smoother and more segment-focused, supporting the fact that   
    updating encoder weights significantly improves the performance (see also ablation in Table \ref{tab:ablation_study}).}
    \label{fig:enc_feats}
\vspace{-0.5cm}
\end{figure}

Input images are first split into a set of fixed size non-overlapping patches before being fed to the encoder, which predicts feature tokens using a ViT-B transformer architecture. 
We formulate the encoder $\mathcal{E}$ as $F, C = \mathcal{E}(I)$
where $I$ are input images and $F$ and $C$ are the feature and class tokens respectively. 
In practice, we use the encoder of a pre-trained MAE~\cite{he2022masked}, a self-supervised vision learner trained to predict masked image patches, which we further finetune on the training dataset to adapt to the different image resolution from ImageNet, on which   MAE~\cite{he2022masked} is trained. As illustrated in Figure ~\ref{fig:enc_feats}, we find that the self-supervised features learnt by the MAE~\cite{he2022masked} encapsulate a powerful intermediate representation of the scene's geometric and appearance properties that can be leveraged by the decoder to predict the radiance field. 
Note that the weights of the encoder are continuously updated during end-to-end training, since we found that this enhances the encoder's ability to generate smoother and more segment-focused features, as shown in Figure ~\ref{fig:enc_feats}, 
and better performance as shown in Table~\ref{tab:ablation_study}.

\begin{figure*}[t]
  \centering
  \includegraphics[width=\textwidth]{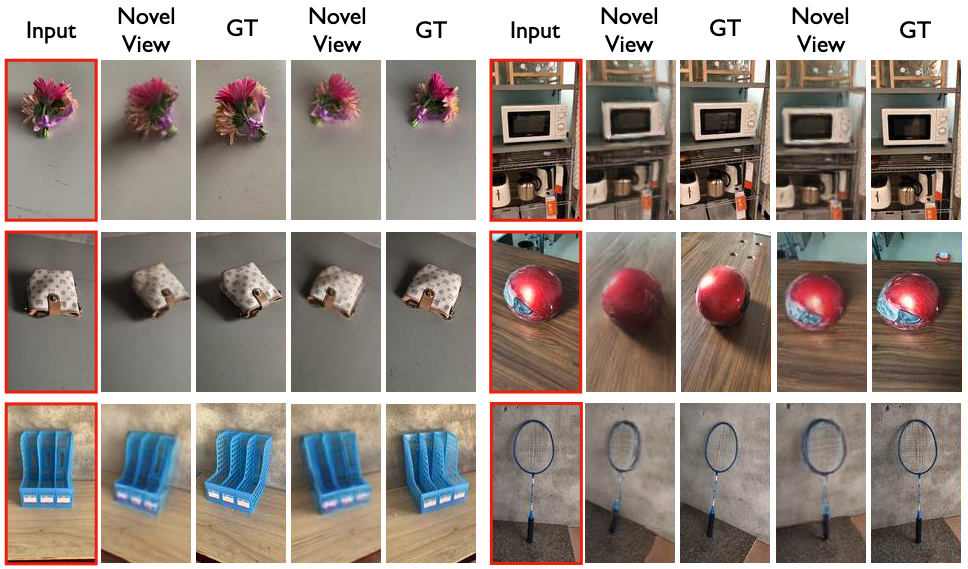}
  \caption{\textbf{Qualitative Results on Test (Unseen) Scenes:} We show the capabilities of \OURS{} to synthesize novel views of unknown scenes. The model correctly synthesizes images from different viewpoints of various categories with diverse backgrounds and scales.}
  \label{fig:novel_view_mvimgnet}
  \vspace{-0.5cm}
\end{figure*}

\subsection{Decoder} 
The goal of the decoder $\mathcal{D}$ is to take the class and feature tokens, $C$ and $F$, predicted by the encoder along with the camera parameters used as conditioning (normalized focal length $f$ and camera distance $z$) and predict the radiance field, which we encode using a vector-matrix (VM) representation, parameterized with three matrices $M$ and three vectors $V$. A key feature of our decoder is the use of cross-attention and self-attention mechanisms.
Unlike previous models~\cite{yu2021pixelnerf} which project spatially aligned ResNet~\cite{he2016deep} features onto target rays, our decoder learns this 2D to 3D correspondence by learning to associate feature tokens with the relevant output tokens through cross-attention.

The decoder $\mathcal{D}$ has output tokens $O$, which are learnable parameters, initialised following a random normal distribution.
For simplicity, from here on we will use the term output tokens $O$ to refer to the concatenation of output tokens and class token $C$, to which we further add positional embeddings~\cite{vaswani2017attention}.  
\begin{equation}
    M, V = \mathcal{D} (F,C,O,f,z).
    \label{eq:decoder}
\end{equation}
The decoder applies cross-attention between feature tokens $F$ and output tokens $O$, and self-attention amongst output tokens $O$. The attention mechanism allows the network to reason about occlusions, where the class token $C$ acts as a global latent vector.
For each attention block, we apply adaptive layer normalisation~\cite{huang2017arbitrary, peebles2022DiT}, instead of standard layer normalisation, to condition on camera parameters.

\subsubsection{Relative Camera Pose}

A key strength of our approach is that it does not require all objects/scenes in the dataset to be brought into alignment by expressing their pose with respect to a canonical reference frame. Instead, we assume the rotation of the camera associated with the input image $I_i$ to be the identity and we express the rotation of any other image $I_j$ as a relative rotation $R_{i \rightarrow j}$.
We assume that the camera location of the input image is 
$T_{i\rightarrow i} = (0,0,z)$, where $z$ is the normalized distance of the input camera from the origin of the world coordinate frame, located at the centroid of the  3D bounding box containing the point cloud, and $T_{i \rightarrow j}={R_i}^T(T_j-T_i)$.
Zero-1-to-3~\cite{liu2023zero} also adopts a relative-pose based approach, however it assumes objects are located in the centre and uses a 3-DoF camera pose model  (radius, elevation and azimuth), while \OURS{} uses a full 6-DoF model for the camera pose.

\subsubsection{Conditioning on Camera Parameters}
There are several ways to apply conditioning such as concatenating camera parameters to output tokens $O$ or feature tokens $F$. 
However, in \OURS{} we apply adaptive layer normalization, as the camera parameters influence the overall scale of the scene.
Conditioning camera parameters improves the model performance as seen in Table~\ref{tab:ablation_study}.

\noindent\textbf{Positional encoding: } We apply the positional encoding from NeRF~\cite{mildenhall2020nerf} for $f$ and $z$, concatenating up to 4-th sine and cosine embeddings with the original values $z$ and $f$ as $M=\oplus_{k=1}^4(\sin 2^k(f), \cos 2^k(f), \sin 2^k(z), \cos 2^k(z))$. 

\noindent\textbf{Adaptive Layer Normalisation: } We employ an additional MLP $\mathcal{A}$ to regress the shift $\delta$, scale $\alpha$ and gate scale $\gamma$ from the conditioning inputs $(z, f, M)$ for each attention block.
\begin{equation}
    \alpha, \delta, \gamma = \mathcal{A}(z, f, M)
    \label{eq:adaptive_layer_norm}
\end{equation}

An alternative to adaptive layer normalisation would be to concatenate $M$ to output tokens $O$, but as seen in our ablation (Table~\ref{tab:ablation_study})  this strategy does not lead to better results.
 
\subsubsection{Attention Blocks}

\OURS{} uses self-attention blocks between output tokens $O$ and cross-attention blocks between output tokens $O$ and feature tokens $F$. The embedding dimension for both $O$ and $F$ is $e$.  
While standard layer normalization is applied to feature tokens $F$, we apply adaptive layer normalization to output tokens $O$ using the shift $\delta$ and scale $\alpha$ values regressed by $\mathcal{A}$, the MLP described in Equation~\ref{eq:adaptive_layer_norm}.

\begin{equation}
    \begin{split}
    O_n &= \delta + \alpha \times \text{Norm}(O) \\
    F_n &= \text{Layer Norm}(F).
    \end{split}
    \label{eq:shift_scale}
\end{equation}
Cross-attention can then be expressed as  
\begin{equation}
    \text{Attn} = \text{Softmax}(\frac{O_n F_n^T}{\sqrt{e}}).
    \label{eq:cross_attention}
\end{equation}
Finally, output tokens are updated via residual connection 
%
 $    O \leftarrow O + \gamma \times \text{Attn}\cdot F_n$, 
where $\gamma$ is the gate scale also regressed by $\mathcal{A}$. Note that self-attention is obtained in an equivalent way, just between output tokens. 

 
\noindent \textbf{Reshaping:} We use MLPs to reshape the output tokens into the vector-matrix representation that encodes the radiance field, adapting their respective dimensionalities. This is followed by unpatchifying into the $3$ matrices and $3$ vectors that form the VM representation. 
Please refer to the supplementary material for more details.

\subsection{Rendering}
The VM representation of the radiance field predicted by the decoder is used to query 3D point features which are then decoded by a multi-layer perception into color $\mathbf{c}$ and density $\mathbf{\sigma}$ and finally rendered via volumetric rendering.

\noindent\textbf{Vector-Matrix Representation:} 
For a compact yet expressive representation of the radiance field, we adopt the vector-matrix decomposition proposed by TensoRF~\cite{chen2022tensorf} which expresses each voxel as the sum of three vectors and matrices, one pair per axis.
Specifically, a 3D feature grid ($\mathcal{T}$), is decomposed into three vectors ($V_{r_1}^X, V_{r_2}^Y, V_{r_3}^Z$) and three matrices ($M_{r_1}^{Y,Z}, M_{r_2}^{Z,X}, M_{r_3}^{X,Y}$), each pair sharing the same channel dimensions ($k$) such that 
\begin{equation}
\mathcal{T} = \sum_{r_1=1}^k V^X_{r_1} \circ M_{r_1}^{Y,Z} + \sum_{r_2=1}^k V^Y_{r_2} \circ M_{r_2}^{Z,X} + \sum_{r_3=1}^k V_{r_3}^Z \circ M_{r_3}^{X,Y}.
\label{eq:vm_representation}
\end{equation}
While the value of density $\mathbf{\sigma}$ is obtained by applying ReLU activation directly to the feature value $\mathcal{T}_{\mathbf{x}}$ at point $\mathbf{x}$, the colour $\mathbf{c}$ is predicted with a shallow MLP, conditioned on the viewing direction. 
The VM representation outperforms using a triplane as shown in our ablation study (Table~\ref{tab:ablation_study}).

\noindent\textbf{Volume Rendering:} 
For each sampled ray $r$, we obtain its final color $\hat{C}(r) \in \mathbb{R}^3$ using volumetric rendering, following the methodology of NeRF~\cite{mildenhall2020nerf}. The transmittance $T_i$ is first computed at each point $\mathbf{x}$ along the ray as $T_i= \exp(-\sum_{j=1}^{i-1}\sigma_j \delta_j)$, where $\delta_i$ is the distance between adjacent points $\delta_i = t_{i+1} - t_i$. The pixel color is calculated by integrating the predicted color at each $i$-th point $\mathbf{c}_i$, weighted by light absorbance $T_i$ - $T_{i+1}$.
\begin{equation}
    \hat{C}(r) = \sum_{i=1}^\text{N} T_i(1-\exp(-\sigma_i\delta_i))\mathbf{c}_i
    \label{eq:rendering}
\vspace{-.1cm}
\end{equation}

\subsection{Training Losses}

We employ a combination of losses to train our architecture in an end-to-end manner including the L2 photometric rendering loss, a Learned Perceptual Image Patch Similarity (LPIPS) loss~\cite{zhang2018unreasonable}, and the distortion-based regulariser proposed by ~\cite{barron2022mip}.  
Given ground-truth pixel colors $\bf{v}$, estimates $\bf{\hat{v}}$ and accumulated transmittance values $\bf{w}$ along the points $\bf{x}$ on the ray, our overall loss function is defined as
\begin{equation}
\mathcal{L}=\mathcal{L}_2(\bf{\hat{v}},\bf{v}) + \lambda\mathcal{L}_\text{LPIPS}(\bf{\hat{v}},\bf{v})  + \beta\mathcal{L}_\text{dist}(\bf{w}, \bf{x}).
\end{equation} 
For MVImgNet $\lambda=0.1$ and $\beta=0.01$. We found LPIPS to be extremely effective on real world datasets (see Table~\ref{tab:ablation_study}).

\section{Experimental Evaluation}

\subsection{MVImgNet} 
\noindent\textbf{Train/Test Split:} MVImgNet contains 
videos of over $6.5$ million real-world scenes across $238$ categories. Our training set, contains a subset of $1.14$M frames across $38$K scenes of $177$ categories. For the test set, we hold out every $100^{th}$ scene from each category to a total of $13,228$ frames, from $447$ scenes and $177$ categories.

\noindent\textbf{Pre-processing:} MVImgNet uses COLMAP to estimate camera matrices and generate 3D point clouds. Since each scene has its own scale from SfM, we rescale point clouds to a unit cube, such that the maximum distance along one axis equals $1$.
Then, we center the point clouds in the world coordinate system. We downsample the original images by $\times12$ while preserving their aspect ratio. Camera intrinsics are recalibrated accordingly.
  
\noindent\textbf{Implementation Details:} \OURS{} has approximately $216$M parameters: $85$M for the encoder, $131$M for the decoder, and $7$K for the renderer. The encoder takes images  of size $160\times90$, using a patch size of $5$, so the total number of feature tokens is $576$. The resolution of the VM representation is $48$. The patch size for the decoder is $3$, the total number of output tokens $816$, and $16$ heads.
The embedding dimension is $768$ for both encoder and decoder, and we sample $48$ points along the ray.
We train \OURS{} with 2$\times$ A100-40GB GPUs for approximately one week, using a batch size of $22$ images and rendering $330$K pixels, up to one million iterations. The initial learning rates are 6e-5 for encoder, and 4e-4 for decoder and renderer and we decay them following the half-cycle cosine schedule. %

\begin{figure}
  \centering
  \includegraphics[width=0.45\textwidth]{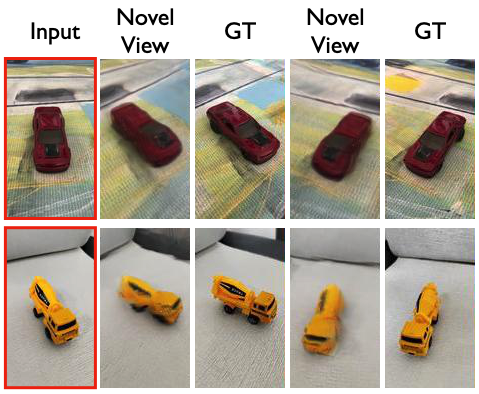}
    \caption{\textbf{Results on Unseen Category:}  This figure shows how the model generalises to a novel category unseen at training. We validate our model with a held-out category (toy-cars).}
    \label{fig:novel_category}
\vspace{-0.2cm}
\end{figure}

\begin{figure}
  \centering
  \includegraphics[width=0.45\textwidth]{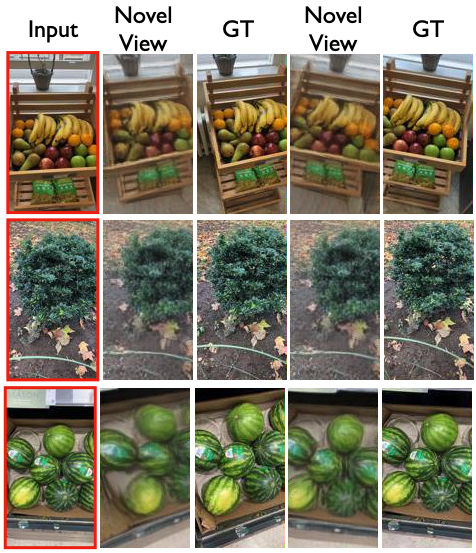}
    \caption{\textbf{Casual Phone Captures:} We demonstrate \OURS{} in OOD  scenarios. First row: the model can capture different categories. Second row: outdoor setting. Third row: many objects.} 
    \label{fig:out_of_distribution}
\vspace{-0.2cm}
\end{figure}

\subsubsection{Qualitative Results}
\noindent\textbf{Results on Test (Unseen) Scenes:} As depicted in Figure \ref{fig:novel_view_mvimgnet}, our model demonstrates its capability to synthesise new views of unknown scenes.
Figure~\ref{fig:novel_view_mvimgnet} highlights the model's ability to handle objects from diverse categories, such as flowers, bags, helmets and others. We show that 
\OURS{} can deal with a variety of backgrounds, such as tables, wooden floors, textureless walls or more cluttered environments.

\noindent\textbf{Results on Unseen Category:} We take a held-out category (toy-cars), and test the ability of our model to generalize to  categories unseen at training. Figure~\ref{fig:novel_category} shows that \OURS{} can synthesize new views of objects from a category not seen at training time given a single input image.

\noindent\textbf{Casual Phone Captures:} Our motivation for this paper was to train a feed-forward model that we could easily use on casually captured images, for instance acquired with our own mobile devices. We test the ability of \OURS{} to deal with out-of-distribution scenes/images by performing zero-shot new-view synthesis on scenes captured by us on a mobile-phone device. We pre-process the images in the same manner as MVImgNet, using COLMAP to estimate the focal length and camera distance parameters to be used as conditioning. 
Figure~\ref{fig:out_of_distribution} shows results on out-of-distribution scenes. 
The top row highlights the model's ability to process a scene with multiple objects from diverse categories. The second row reveals its competence on outdoor scenes, despite their limited presence in the training set.
The third row illustrates \OURS{}'s ability to learn scenes with a large number of  objects. Figure~\ref{fig:teaser} shows two further examples of results on phone captures. 

\begin{figure}
  \centering
  \includegraphics[width=0.45\textwidth]{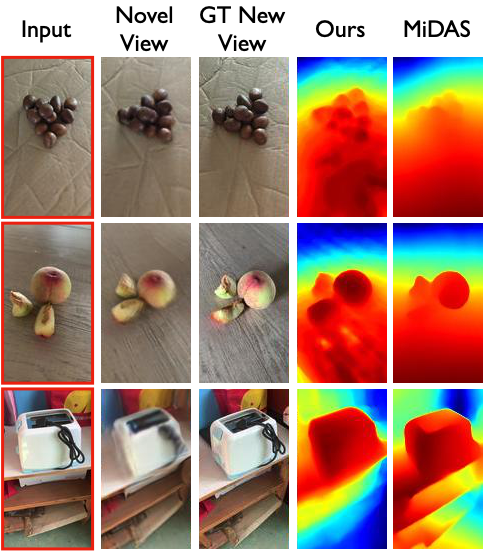}
    \caption{\textbf{Depth Estimation:} Examples of the depth estimates on test images. Although \OURS{} focuses on novel view synthesis and is trained with RGB losses only, depth estimation is consistent and finds good object boundaries.
    We show a comparison with the state-of-the-art disparity estimator MiDAS v3.1 with Swin2-L384~\cite{birkl2023midas}. We provide MiDAS the GT new view images as input, as it cannot do novel view synthesis. \OURS{} performs well even though it is not trained with depth supervision, unlike MiDAS.}
    \label{fig:depth}
\vspace{-0.3cm}
\end{figure}

\noindent\textbf{Depth Estimation:} Figure ~\ref{fig:depth} shows qualitative results of the depth predicted by \OURS{}.
Since there is no ground-truth depth for these images, we qualitatively compare ~\OURS{} with the recent MiDASv3.1 with Swin2-L384~\cite{birkl2023midas}.
As MiDAS predicts depth from RGB images, we provide the GT novel view image as input.
This shows that depth estimation with \OURS{} performs well even though it is not trained with depth supervision, while MiDAS uses direct depth supervision from multiple datasets.

\begin{figure}[ht]
  \centering
  \includegraphics[width=0.5\textwidth]{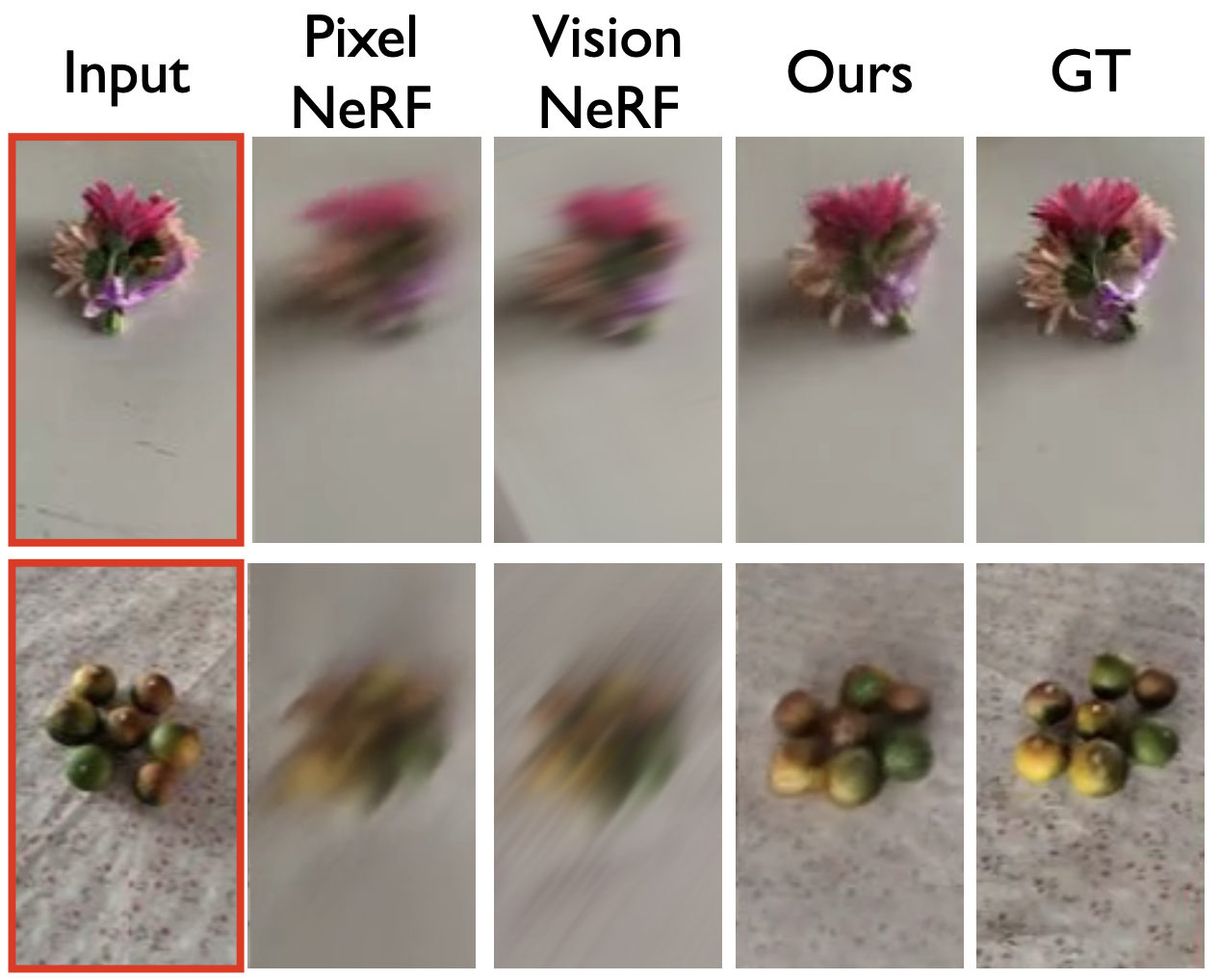}
    \caption{\textbf{Qualitative Comparison with PixelNeRF and VisionNeRF:} \OURS{} displays better performance than PixelNeRF~\cite{yu2021pixelnerf} and VisionNeRF~\cite{lin2023visionnerf}, especially when the target view is far away from the input view.}
\label{fig:compare_with_pixelnerf}
\vspace{-0.2cm}
\end{figure}

\begin{table}
    \centering
    \begin{tabular}{@{}llll@{}}
         &  \multicolumn{1}{c}{PSNR$\uparrow$} & \multicolumn{1}{c}{SSIM$\uparrow$} & \multicolumn{1}{c}{LPIPS$\downarrow$} \\ \midrule
        PixelNeRF~\cite{yu2021pixelnerf} & 17.02 & 0.41 & 0.54 \\
        VisionNeRF~\cite{lin2023visionnerf} & 19.82  & 0.51 & 0.47 \\
        Ours & \textbf{20.83} & \textbf{0.57} & \textbf{0.29}  \\ 
        \bottomrule
    \end{tabular}
    \caption{\textbf{Quantitative Tests on MVImgNet:} We compare \OURS{} with PixelNeRF and VisionNeRF on new view synthesis from single in-the-wild images. \OURS{} outperforms both on all metrics.}
    \label{tab:quantitative_results_mvimgnet}
\end{table}

\subsubsection{Comparisons with baseline models}

\begin{figure}
    \centering
    \includegraphics[width=0.5\textwidth]{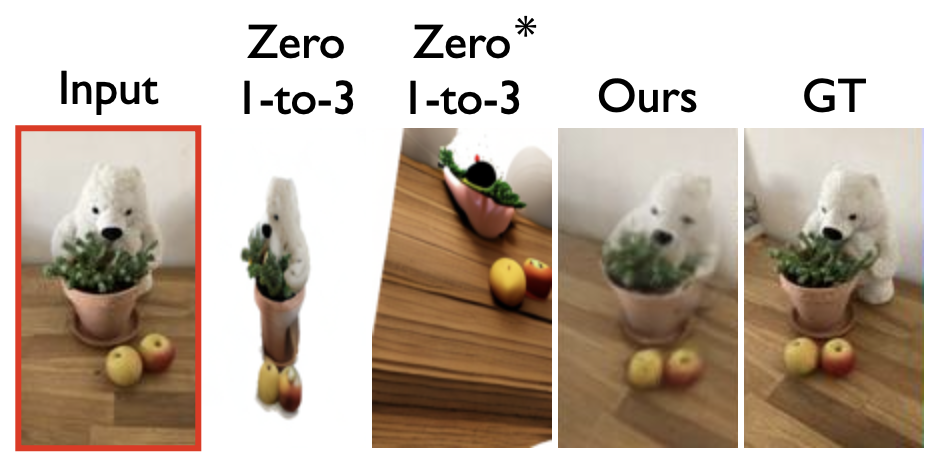}
    \caption{\textbf{Qualitative comparison with Zero-1-to-3~\cite{liu2023zero} on a phone captured input image:}   Zero-1-to-3 shows result w/masked input, Zero1to3$^{*}$ shows result w/full image input. None result in good novel-view synthesis, except \OURS{}.}
    \label{fig:compare_with_zero123}
\end{figure}

We compare \OURS{} with PixelNeRF~\cite{yu2021pixelnerf} and VisionNeRF~\cite{lin2023visionnerf}, all trained on MVImgNet using their official code releases and applying the same pre-processing as~\OURS{}.
Figure~\ref{fig:compare_with_pixelnerf} and Table~\ref{tab:quantitative_results_mvimgnet} show that ~\OURS{} outperforms both models by a large margin. Both models rely on aligned features, but have limitations dealing with occlusion.
We qualitatively compare with the pre-trained Zero-1-to-3~\cite{liu2023zero} on a phone capture in Figure ~\ref{fig:compare_with_zero123}, using the approximate camera pose as it assumes a centered object. Since Zero-1-to-3 and its follow-up works assume a 3DoF camera and do not model the background, we could not conduct quantitative comparisons. 

We could not conduct quantitative comparisons with generative models such as GenVS~\cite{chan2023generative} or with the large-scale model LRM~\cite{hong2023lrm} as their models are not publicly available. 
Moreover, GenVS is category-specific and LRM does not model backgrounds.

\begin{table}[bt]
    \centering
    \begin{tabular}{@{}lccc@{}}
         &  \multicolumn{1}{c}{PSNR$\uparrow$} & \multicolumn{1}{c}{SSIM$\uparrow$} & \multicolumn{1}{c}{LPIPS$\downarrow$} \\ \midrule
        Ours & \textbf{20.83} & \textbf{0.57} & \textbf{0.29}  \\ 
        w/o LPIPS & 20.72 & 0.54 & 0.46  \\ 
        w/o Camera Conditioning  & 20.24 & 0.49 & 0.36  \\
        Concat Camera Parameters\textsuperscript{(1)} &  20.81 & \textbf{0.57}  &  0.30\\
        w/ Self-Attention Decoder\textsuperscript{(2)} & 20.74 & \textbf{0.57} & 0.31\\
        w/o VM Representation\textsuperscript{(3)} & 19.60 & 0.49 & 0.44 \\
        w/o Updating Encoder & 18.54 & 0.47  & 0.49 \\
        \bottomrule
    \end{tabular}
    \caption{\textbf{Ablation:} 
    For (1) we concatenate high dimensional camera feature tokens to output tokens instead of adaptive layer normalisation. For (2) we update all tokens via self-attn (no cross-attn). For (3), we use triplane instead of VM as the representation.}
    \label{tab:ablation_study}
\end{table}

\subsubsection{Ablation Study}
We conducted an ablation study to analyse the effect of design choices on the performance of ~\OURS{} as summarised in Table~\ref{tab:ablation_study}.
While the perceptual LPIPS loss~\cite{zhang2018unreasonable} is not commonly employed in NeRF-based methods, it appears to play a crucial role in improving quality.
Conditioning on normalized focal length and camera distance helps the model deal with scale ambiguities, and adaptive layer normalisation performs better than concatenating camera parameters to output tokens.
Instead of employing cross-attention in Figure~\ref{fig:overall}, we can concatenate feature tokens and output tokens and update both of them using self-attention. However, it increases memory consumption and does not lead to a better result in Table~\ref{tab:ablation_study}.
While projecting features onto triplanes has been extensively used before~\cite{chan2022efficient, sargent2023vq3d,le2023stylemorph, hong2023lrm}, our experiments show that the use of a vector-matrix (VM) representation~\cite{chen2022tensorf} improves performance.
Note that for the triplane representation, we use a 1-layer MLP to regress occupancy, while occupancy is directly calculated using Equation~\ref{eq:vm_representation} in our VM representation.
Updating encoder weights also improves the performance as the encoder output is used for cross-attention.

\begin{table}
    \centering
        \label{tab:quantitative_results}
        \begin{tabular}{@{}lllll@{}}
            \toprule
            & \multicolumn{2}{c}{Cars} &  \multicolumn{2}{c}{Chairs}                                 \\ \cmidrule(lr){2-3} \cmidrule(l){4-5} 
             &  \multicolumn{1}{c}{PSNR$\uparrow$} & \multicolumn{1}{c}{LPIPS$\downarrow$} & \multicolumn{1}{c}{PSNR$\uparrow$} & \multicolumn{1}{c}{LPIPS$\downarrow$}   \\ \midrule
            PixelNeRF~\cite{yu2021pixelnerf} & 23.17 & 0.146 & 23.72 & 0.128 \\
            VisionNeRF~\cite{lin2023visionnerf} & 22.88 & 0.084 & 24.48 & 0.077  \\
            VD~\cite{szymanowicz23viewset_diffusion}  & 23.29 & 0.094 & - & - \\ 
            SSDNeRF~\cite{chen2023single} & 23.52 & \textbf{0.078} & 24.35 & \textbf{0.067}\\
            \midrule
            Ours & \textbf{23.91}& 0.122 & \textbf{24.50} & 0.090  \\ 
            \bottomrule
        \end{tabular}
        \caption{\textbf{Quantitative Evaluation on ShapeNet-SRN:} \OURS{} performs similarly to other baselines on ShapeNet Cars/Chairs, although our method only requires relative pose. For qualitative comparisons, see supplementary materials.}
    \label{tab:shapenetsrn}
\vspace{-0.3cm}
\end{table}

\subsection{ShapeNet-SRN}

ShapeNet-SRN~\cite{sitzmann2019scene} has two categories (cars and chairs) and is a widely used benchmark to compare models that perform novel view synthesis from a single input image. 
Since all objects are aligned and there is no scale ambiguity, pre-processing is not needed and we do not use the LPIPS loss as it is a synthetic dataset. Table ~\ref{tab:shapenetsrn} shows that \OURS{} performs similarly to baseline models on ShapeNet-SRN dataset.
Although other models apply absolute pose, we only employ relative pose, which means we do not fully exploit the alignment of objects in ShapeNet-SRN.
 
\section{Conclusion}

We have introduced \OURS{}, a transformer-based scalable model for new view synthesis from a single in-the-wild image.
Our evaluations demonstrate robust performance on the MVImgNet test set, novel category synthesis and phone captures of out-of-distribution scenes.
Our design choices were validated via ablations and a quantitative comparison was conducted on MVImgNet and ShapeNet-SRN. 
Interesting future directions include extending \OURS{} to adopt a probabilistic approach and to multiview inputs.

\noindent\textbf{Limitations: } Some loss of sharpness could be due to our computational constraints, which led us to downsample images by $\times12$ and train on a fraction of the original dataset. 
We pushed for a transformer-based architecture, without GAN losses or SDS~\cite{poole2022dreamfusion}, which eased and sped up training, but may have also contributed to some loss of detail.

\noindent\textbf{Acknowledgements: } 
Our research has been partly supported by a sponsored research award from Cisco Research and has made use of time on HPC Tier 2 facilities Baskerville (funded by EPSRC EP/T022221/1 and operated by ARC at the University of Birmingham) and JADE2 (funded by EPSRC EP/T022205/1). We are grateful to N.~Mitra and D.~Stoyanov for fruitful discussions.

{
    \small
    \bibliographystyle{ieeenat_fullname}
    \bibliography{main}
}

\clearpage
\maketitlesupplementary

\setcounter{section}{0}
\renewcommand{\thesection}{\Alph{section}}

\section{Implementation Details}

\noindent\textbf{Finetuning MAE Encoder:} We use the pre-trained MAE~\cite{he2022masked} with ViT-B~\cite{dosovitskiy2020image} from the original MAE implementation. 
Those weights are trained for ImageNet~\cite{deng2009imagenet} which has a resolution of $224\times 224$ pixels with a patch size $16$.
This means that the model divides the image into $196$ feature tokens.
Our image resolution for MVImgNet~\cite{yu2023mvimgnet} is $160 \times 90$, and we use an encoder patch size of $5$, resulting in $576$ patches in the encoder.
During fine-tuning, we initialise the weights of attention blocks with the pre-trained MAE, as the Transformer architecture allows for arbitrary attention matrix shapes as long as the embedding dimension remains the same.
We fine-tune by randomly masking out and inpainting patches with L2 reconstruction loss,similar to the approach used in MAE~\cite{he2022masked}.
The process converges within a single epoch.

\noindent\textbf{Initialisation of Decoder:} We initialise the decoder of \OURS{} with the fine-tuned MAE weights.
With the exception of the learnable parameters of positional embedding of output tokens and the last MLP layers, we initialise the weights of attention blocks with the fine-tuned MAE weights.

\noindent\textbf{Number of output tokens:} For MVImgNet~\cite{yu2023mvimgnet}, the resolution of vector-matrix(VM) representation is $48$, and the channel dimension of each matrix and vector is $32$.
The patch size of the decoder is $3$.
Each $48\times48$ matrix $M$ consists of non-overlapping $16\times 16$ patches, and the $48$ dimensional vector $V$ is divided into $16$ patches.
Therefore, the total number of output tokens for VM representation is $818$.

\noindent\textbf{Decoder MLPs and Reshaping:} The embedding dimension of the decoder is $768$. 
We have $818$ output tokens, and the channel dimension of VM representation~\cite{chen2022tensorf} is $32$, with a patch size of $3$ for the decoder.
For the output tokens corresponding to the matrices $M$ in the VM representation, we deploy MLP to reduce the embedding dimension to $288$.
For those corresponding to vectors V, we reduce it to $96$.
Subsequently, we reshape them into VM representation.

\begin{figure}
  \centering
  \includegraphics[width=0.45\textwidth]{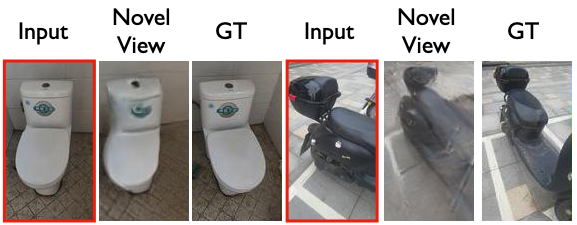}
    \caption{\textbf{Failure Cases} This figure illustrates when the model fails to do new view synthesis properly. The toilet scene shows that the model learns geometry in a distorted way. In the motorcycle scene, the model fails to estimate the occluded area and the proper scale.}
    \label{fig:failure}
\vspace{-0.3cm}
\end{figure}

\section{Qualitative Results on ShapeNet-SRN}

\begin{figure*}[t]
  \centering
  \includegraphics[width=0.8\textwidth]{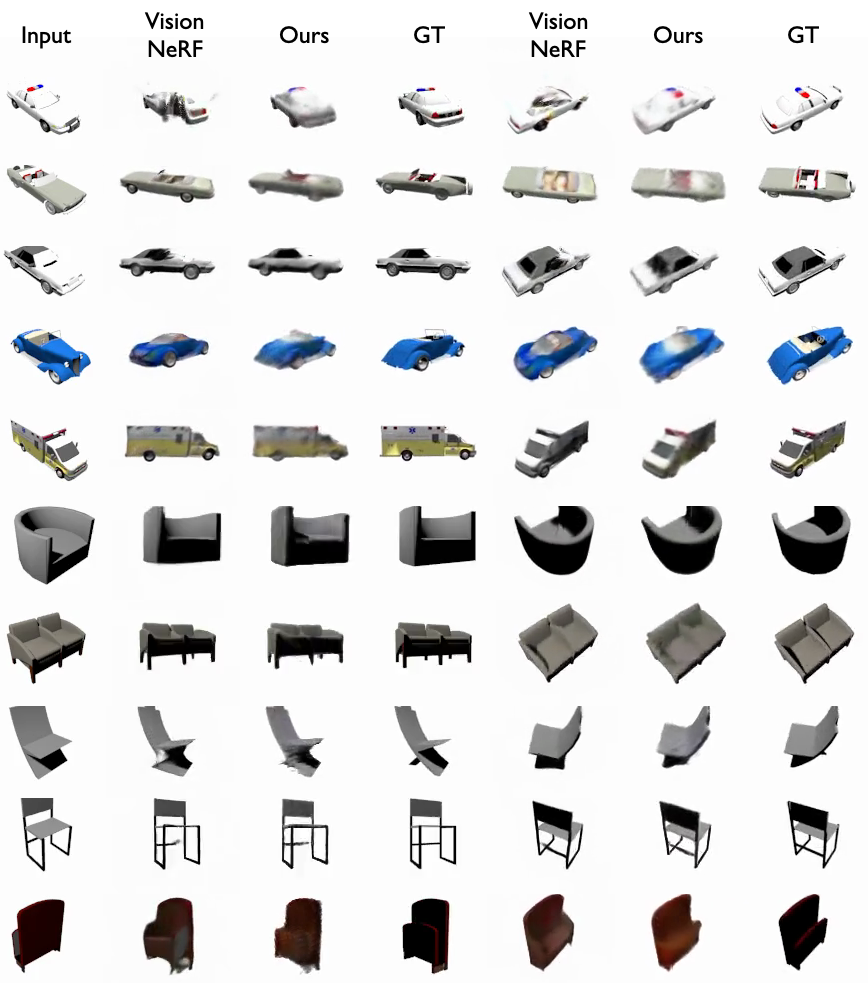}
    \caption{\textbf{Qualitative Comparison on ShapeNet-SRN~\cite{sitzmann2019scene}: }\OURS{} performs similar to VisionNeRF which is one of the top-performing models on ShapeNet-SRN dataset. Note that we do not employ LPIPS and do not condition on camera parameters for ShapeNet-SRN as it is a synthetic dataset, but we still use the relative pose even though objects are aligned in 3D. }
    \label{fig:compare_with_shapenet}
\vspace{-0.5cm}

\end{figure*}

We perform a qualitative comparison with VisionNeRF~\cite{lin2023visionnerf} on ShapeNet-SRN~\cite{sitzmann2019scene} dataset as depicted in Figure ~\ref{fig:compare_with_shapenet}. 
VisionNeRF, recognised as one of top-performing models on ShapeNet-SRN, employs ViT~\cite{dosovitskiy2020image} as its encoder.
Notably, VisionNeRF does not utilise any generative approaches, and was trained using $8$ A100 GPUs.
Similarly for MVImgNet, we fine-tune a MAE for the ShapeNet-SRN dataset and initialise the parameters of both encoder and decoder of \OURS{} with this fine-tuned MAE for ShapeNet-SRN.
The ShapeNet-SRN images are of resolution $128\times 128$, and we use an encoding patch size of $8$, resulting in $256$ feature tokens.
The resolution of VM representation is $64$, and the decoder patch size is $4$, so we use $818$ output tokens, each with an embedding dimension of the Transformer as $768$. 
We still maintain the relative pose but do not condition on camera parameters as the dataset is aligned and does not have scale ambiguities.
We train the model with a single $3090$ GPU with $500,000$ and $700,000$ iterations, respectively for car and chair.

\begin{figure*}[t]
  \centering
  \includegraphics[width=\textwidth]{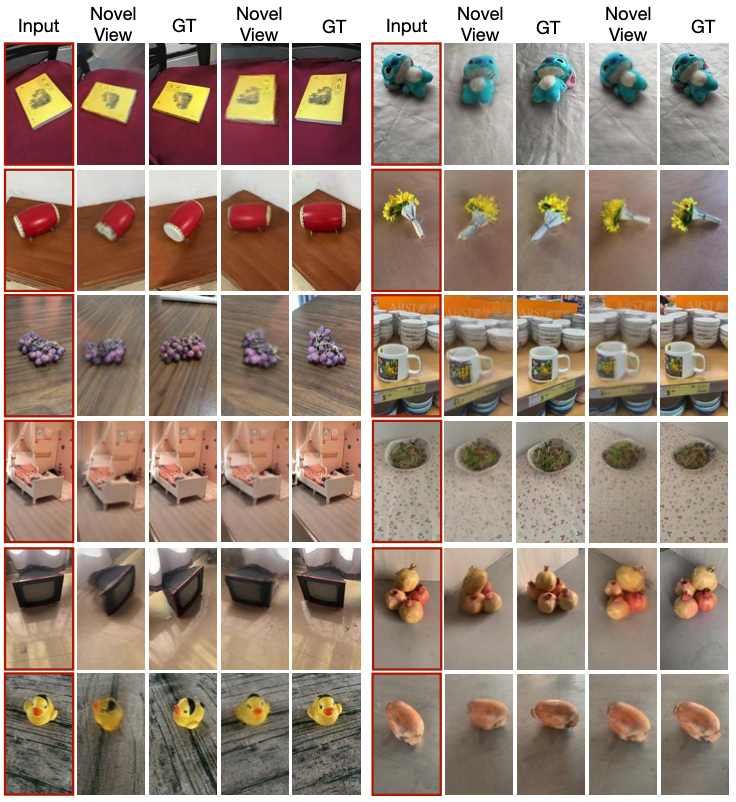}
    \caption{\textbf{Qualitative Results on Test (Unseen) Scenes of MVImgNet~\cite{yu2023mvimgnet}: } \OURS{} can synthesize high-quality novel view on challenging scenes from single in-the-wild input images.}
    \label{fig:compare_mvimgnet}
\vspace{-0.5cm}

\end{figure*}

\end{document}